\documentclass{article}
\usepackage{times}
\usepackage{proceedings}
\usepackage{latexsym}
\usepackage{amssymb}
\newtheorem{theorem}{\bf Theorem}

\newtheorem{corollary}{\bf Corollary}
\newtheorem{observation}{\bf Observation}

\newtheorem{definition}{\bf Definition}

\newcommand{\qed}{\rule{2mm}{2mm}}
\newcommand{\mdot}{\stackrel{.}{-}}

\newenvironment{proof}{\noindent {\bf Proof:}}{\hfill \qed\\}
\title{Iterated revision and the axiom of recovery: a unified treatment via epistemic states}
\author{
{\bf Samir Chopra}\\
schopra@cse.unsw.edu.au\\
Knowledge Systems Group \\
University of New South Wales \\
Sydney, NSW 2052, Australia \\
\And
{\bf Aditya Ghose}\\
aditya@uow.edu.au\\
Department of Information Systems \\
University of Wollongong \\
Wollongong, NSW 2522, Australia \\
\And
{\bf Thomas Meyer} \\
tmeyer@cs.up.ac.za\\
Computer Science Department\\
University of Pretoria\\
Pretoria 0002, South Africa\\
}
\begin{document}
\maketitle
\begin{abstract} The axiom of recovery, while capturing a central intuition
regarding belief change, has been the source of much controversy. We
argue briefly against putative  counterexamples to the axiom---while agreeing that some
of their insight deserves to be preserved---and present additional recovery-like axioms in a
framework that uses epistemic states, which encode preferences,  as the object of revisions. This provides
a framework in which iterated revision becomes possible and makes
explicit the connection between iterated belief change and the axiom of
 recovery. We
provide a representation theorem that connects the semantic conditions
that we impose on iterated revision and the additional syntactical
properties mentioned. We also show some interesting similarities between our
framework and that of Darwiche-Pearl \cite{Darwiche-ea:97a}.
In particular, we show that the
intuitions underlying the controversial (C2) postulate are captured
by the recovery axiom and our recovery-like postulates (the latter can be seen as weakenings of (C2). \end{abstract}

\section{Introduction}

A particularly simple sequence of belief change in reasoning agents is
that of giving up and then adopting the same belief (``I believed I had
money for the movies, but then realized I had left my wallet at home. However, a
few minutes later, I discovered a twenty in my pocket and regained my belief that I had enough money for the movies''). The axiom of
recovery in the AGM framework \cite{agm85} attempts to place a rationality constraint
on the form of such a change. It states that expansion by a belief
recovers any beliefs lost by the previous contraction by that belief.  The status of the
axiom of recovery has been a source of much controversy in belief revision
\cite{Fuhrmann:90a,Hansson:91a,Hansson:93c,Levi:91a}.  There are
well-known counterexamples to recovery, with the most convincing ones amongst these being Hansson's
Cleopatra and George-the-criminal examples \cite{Hansson:91a,Hansson:98a}.
The following is a slightly amended version of the former:

\begin{quote} I believe that `Cleopatra had a son' ($\phi$) and that
`Cleopatra had a daughter' ($\psi$), and thus also that `Cleopatra had a
child' ($\phi \vee \psi$). Then I receive information that Cleopatra had
no children, which makes me give up my belief in $\phi \vee \psi$. But
then I am told that Cleopatra did have children, and so I add $\phi \vee
\psi$. But I should not regain my belief in either $\psi$ or $\phi$ as a
result.  \end{quote}

One response to this situation is to isolate a class of belief change operators
that do not satisfy recovery i.e., the so-called withdrawal operators
\cite{Makinson:87a}.
We do not adopt this approach for a couple of reasons. Firstly, withdrawal
operators violate the principle of minimal change \cite{Hansson:98a}. As
an example, consider the operator $\mdot$ defined as follows ($K$ is a
belief set closed under logical consequence, $\alpha$ an arbitrary epistemic input): if $\alpha
\not\in K$, then $K \mdot \alpha = K$, otherwise, $K \mdot \alpha =
Cn(\emptyset)$. A fundamental intuition behind minimal contraction, the
principle of core-retainment\footnote{The principle of core-retainment
states that if $\beta \in K$ and $\beta \not\in K \mdot \alpha$ then there
is a set $K'$ such that $K' \subseteq K$ and that $\alpha \not\in Cn(K')$
but $\alpha \in Cn(K' \cup \{\beta\})$; it requires of an excluded
sentence $\beta$ that it in some way contribute to the implication of
$\alpha$ from $K$.}, is only satisfied by withdrawal operators if they
satisfy the recovery axiom as well. This should reinstate our faith in
the recovery axiom since it is hard to find a satisfactory alternative formalization of
the intuition that beliefs that do not contribute to $K$ implying $\alpha$
should be retained in $K \mdot \alpha$.
So, while the counterexamples do tickle our intuitions, it is equally the case that there is an important intuition about rational
belief change that the recovery postulate captures.
Indeed, the recovery postulate is best thought of as a
version of the principle of minimal change: so much
of the original belief state is retained on contraction that the original
belief state can simply be restored on adopting the same belief. Our
approach to this situation is that even if the original postulate is
rejected as being too permissive, {\em some} recovery like postulates must
constrain belief revision if the principle of minimal change is to be
taken seriously.
Furthermore,
recovery follows from other highly plausible postulates such as closure, inclusion, vacuity, success, extensionality and core-retainment \cite{Hansson:98a}.  Significantly, there is a clear
and intimate connection between iterated revision and the recovery axiom:
we can view the axiom as specifying the form
of the iterated revision that should take place when contracting and
revising by the same belief. In what follows, we will make this connection
clearer.


But what about the counterexamples? Surely, they point to counterintuitive
scenarios arising from the adoption of the recovery axiom? We argue that, underlying these examples is an assumption that
information leading to the specified sequence of contraction and expansion is not received from the same source. That is, our claim is
that recovery should always hold when restricted to the case where
information is obtained from the same source, but that it need not hold
when information is obtained from different sources.
Consider the Cleopatra counterexample. The agent believes both $\phi$ and
$\psi$ originally, and as a result is committed to the belief that
$\phi \vee \psi$. Now the agent receives information to the effect that
$\neg (\phi \vee \psi)$. Crucially, what is left out of this example is
details about the sources of the epistemic inputs.
If source $S_1$ provides the {\em reasons} for believing
$\neg (\phi \vee \psi)$ and source $S_2$ provides the reason for believing
$\phi \vee \psi$ then it makes
sense to think that the agent does not recover its original beliefs
in $\phi$ or $\psi$. However, if it is the same source that provides
information on both $\neg (\phi \vee \psi)$ and $\phi \vee \psi$, then why
should the agent not regain
its belief in $\phi$ and $\psi$? After all, source $S_1$ provided
the reason for the agent dropping its belief in $\phi$ and $\psi$ in the first place. If it then
supplies information to the contrary, the agent's reasons for dropping those
beliefs have been negated, and it should regain its original beliefs. To
do otherwise would be counterintuitive.
If however it is another source
that provides the new information, then the agent's original reasons for
contracting by $\phi$ and $\psi$ remain unaffected
and there is no reason for it to start believing $\phi$ or $\psi$ again. (For a similar though crucially different response see \cite {nay93}).

The issue of what happens when information is
obtained from different sources is interesting in its own right, and
deserves to be treated separately. In general our attitude is that the
intuitions behind the recovery axiom are worth capturing:
it attempts
to place rational constraints on what happens when we revise and contract by the
same formula. This sort of belief change is commonplace and must be handled by
any adequate formal framework.

\subsection{Our Proposal}

We will consider versions of postulates in the same spirit as recovery. We
argue that to do so, a shift to belief change on epistemic states, in the
Darwiche-Pearl spirit is necessary, since we need a framework in which to
talk about iterated revision. Cantwell \cite{Cantwell:99a} also provides
recovery-like properties in the context of iterated revision, but these
however restate recovery itself in terms of revision (where contracting
with $\alpha$ is replaced by a revision with $\lnot\alpha$). This is done
to show that the counterexamples to recovery are not only a criticism of
AGM contraction (as has been argued in the past), but also a criticism of
AGM revision. Cantwell goes on to show that examples similar to the Cleopatra
and
George-the-criminal examples can be constructed for iterated revision as well.

While adopting the representational framework of epistemic states, we do not accept all the Darwiche-Pearl postulates. There is sufficient
debate in the
belief revision literature on the appropriateness of these postulates.  In principle, though, we are of the opinion
that the 3rd and 4th Darwiche-Pearl postulates are valid. Like others
we feel that the 2nd postulate is too strong. The results in this
paper provide a weaker, and, we think, acceptable alternative to the 2nd
postulate.  We are also of the opinion that the 1st Darwich-Pearl postulate is too
strong (\cite{Meyer:99c} provides examples to back up this claim).
We
adopt the basic setting in which belief change is performed on
epistemic states, from which a total preorder on valuations and a
knowledge base can be extracted.
We provide a set of reformulated
AGM postulates for belief change on epistemic states and \emph{insist} on these.

We present some recovery-like postulates, as well as restrictions on
the way in which the orderings extracted from epistemic states may be
modified when revision and contraction take place, and provide a
representation theorem that connects the recovery-like postulates and the
postulates on orderings.
It turns out that the recovery-like postulates, when combined, can be
thought of as a weakened version of the (C2) postulate of Darwiche-Pearl.
This is brought out clearly when the postulates on orderings are
considered. The link between recovery and the (C2) postulate is
interesting and surprising. This makes it possible to think
of (C2) as having overstated the case and of the recovery postulate and our
weakenings as
having addressed its problems.

\subsection{Notation and basic definitions}

We assume a finitely generated propositional language $L$ closed under the
usual propositional connectives and equipped with a classical
model-theoretic semantics; the constants $\top, \perp$ are in $L$. $V$ is
the set of valuations of $L$ and $M(\alpha)$ is the set of models of
$\alpha\in L$. Classical entailment is denoted by $\models$. Roman letters, $p, q, r, \ldots$ denote propositional atoms; Greek letters $\alpha, \beta, \ldots$ stand for arbitrary formulas. We reserve the letter $\Phi$ to denote epistemic states.  $M_{\preceq_{\Phi}}(\alpha)$ denotes the minimal models of $\alpha$ in the total preorder on valuations associated with the epistemic state $\Phi$.


\begin{definition} Associated with an \emph{epistemic state} $\Phi$ is a total preorder on valuations $\preceq_{\Phi}$, and a knowledge base
$K(\Phi)$.  $M_{\preceq_{\Phi}}(\alpha)$ denotes the minimal models of $\alpha$ in the total preorder on valuations.
The knowledge base associated with the epistemic state is obtained by considering the minimal models
in $\preceq_{\Phi}$ i.e., $M(K(\Phi)) = M_{\preceq_\Phi}(\top)$.
\end{definition}

%

$\phi$ represents the set of all wffs entailed by $\phi$ (the theory
obtained from the set of minimal models in $\preceq_{\Phi}$). Observe that the
knowledge bases extracted from $\Phi$ are all logically equivalent. We
will often abuse notation by using $K(\Phi)$ to refer to the \emph{the}
knowledge base extracted from $\Phi$. The intention is that $K(\Phi)$ is
some canonical representative of \emph{all} the knowledge bases extracted
from $\Phi$.

\subsection{The reformulated AGM postulates}
In the reformulated postulates below, $*$ and $\mdot$ are belief change
operations on epistemic states, not knowledge bases. So $*$ takes an
epistemic state and a sentence and produces an epistemic state. For
$\mdot$ and $*$ to satisfy the AGM postulates means that they satisfy the
reformulated AGM postulates which apply to epistemic states, not knowledge
bases. The reformulated AGM postulates
guarantee a unique extracted knowledge base when revision or contraction
is performed i.e., the lowest level of valuations in the resulting
epistemic state is fixed. What is not fixed is how to order the remaining
valuations. Note that the object of revision is the epistemic state, but
in stating the postulates we specify the form of the knowledge base
extracted from the epistemic state.  Here are the reformulated AGM
postulates. First contraction:

\begin{itemize}
\item ($\Phi-$1):  $K(\Phi-\alpha) =Cn(K(\Phi-\alpha))$
\item ($\Phi-$2):  $K(\Phi-\alpha) \subseteq K(\Phi)$
\item ($\Phi-$3):  If $\alpha \notin K(\Phi)$ then $K(\Phi-\alpha) =K(\Phi)$
\item ($\Phi-$4):  If $\not\models\alpha $ then $\alpha \notin K(\Phi-\alpha)$
\item ($\Phi-$5):  If $\alpha \equiv \beta$ then $\Phi-\alpha =\Phi-\beta$
\item ($\Phi-$6):  If $\alpha \in K(\Phi)$ then $K(\Phi)\subseteq(K(\Phi-\alpha))+\alpha$
\item ($\Phi-$7):  $K(\Phi-\alpha)\cap K(\Phi-\beta)\subseteq K(\Phi-(\alpha\wedge \beta ))$
\item ($\Phi-$8):  If $\beta\notin K(\Phi-(\alpha \wedge \beta ))$ then $K(\Phi-(\alpha \wedge\beta))\subseteq K(\Phi)-\beta$
\end{itemize}
In what follows, we will be particularly interested in the relationship between $K(\Phi *
\alpha - \alpha)$ and $K(\Phi)$. We will show that equality between the two
sides conflicts with the reformulated AGM postulates but does hold
under some conditions.

The intuitions corresponding to the postulates are roughly the same as
those underlying the original AGM postulates. For example, ($\Phi-$1) states
that the knowledge base associated with the revised epistemic state is
closed under logical consequence. ($\Phi-$5) states that contracting by logically equivalent
formulas results in the same epistemic state.  This particular postulate
highlights a difference between the original AGM postulates and
our reformulations. The original AGM postulate requires the belief set after
revision to be the same after revisions by logically equivalent formulas, whereas we require that if two epistemic states are the same, then revisions by logically equivalent formulas should result in the same epistemic state. This is crucially different from merely requiring that the knowledge base associated with
the epistemic state be the same (such a reformulation of the original AGM
axioms by Darwiche-Pearl is responsible for making (C2) compatible with them). Note that we include the recovery axiom above.

The following are the reformulatedreformulated  AGM postulates for revision:

\begin{itemize}
\item ($\Phi*$1) $K(\Phi*\alpha) =Cn(K(\Phi*\alpha))$
\item ($\Phi*$2) $\alpha \in K(\Phi*\alpha)$
\item ($\Phi*$3) $K(\Phi*\alpha)\subseteq K(\Phi)+\alpha $
\item ($\Phi*$4) If $\lnot\alpha \notin K(\Phi)$ then $K(\Phi)+\alpha\subseteq K(\Phi*\alpha)$
\item ($\Phi*$5) If $\alpha \equiv \beta $ then $\Phi*\alpha =\Phi*\beta$
\item ($\Phi*$6) $\bot \in K(\Phi*\alpha)$ iff $\models \lnot \alpha $
\item ($\Phi*$7) $K(\Phi*(\alpha \wedge \beta ))\subseteq K(\Phi*\alpha)+\beta $
\item ($\Phi*$8) If $\lnot \beta \notin K(\Phi*\alpha)$ then $K(\Phi*\alpha)+\beta\subseteq K(\Phi*(\alpha \wedge \beta))$
\end{itemize}

As with the contraction postulates, the intuitions corresponding to the postulates are roughly the same as
those underlying the original AGM postulates. For example, ($\Phi*$1) states
that the knowledge base associated with the revised epistemic state is
closed. ($\Phi*$6) states that an inconsistent knowledge base only results when
revising by contradictions (note the modified ($\Phi*5$) postulate as well).

For the sake of completeness, we include the Darwiche-Pearl postulates for iterated revision \cite{Darwiche-ea:97a} reformulated for our framework. In the
four postulates below
$\circ$ is the update operator, $\alpha, \mu$, represent new epistemic
inputs and $\Phi$ represents an epistemic state.

\begin{description}

\item  (C1) If $\alpha \models \mu$, then $K(\Phi  \circ \mu \circ
\alpha) = K(\Phi \circ \alpha)$.

\item (C2) If $\alpha \models \neg\mu$, then $K(\Phi \circ \mu \circ
\alpha) = K(\Phi \circ \alpha)$.

\item (C3) If $K(\Phi \circ \alpha) \models \mu$, then $K(\Phi \circ
\mu \circ \alpha) \models \mu$.

\item (C4) If $K(\Phi \circ \alpha) \not\models \neg\mu$, then $K(\Phi
\circ \mu \circ \alpha) \models \mu$.  \end{description}

The postulate (C1) is a more powerful version of the ($\Phi \ast 7$) and
($\Phi \ast 8$) postulates (it implies them); it states that when two pieces of
information (one more specific than the other) arrive, the first is made
redundant by the second.  (C2) says that when two contradictory
epistemic inputs arrive, the second one prevails;  the second evidence
alone yields the same belief state. Here the {\em prima facie} connection with
recovery should be obvious; for the basic form of the recovery
axioms deals with `contract by $\alpha$ and then expand by $\alpha$' while
(C2) deals with `revise by $\alpha$ and then revise by (effectively) $\neg\alpha$'. The latter is clearly stronger.
(C3)
says that a piece of evidence $\mu$ should be retained after
accommodating more recent evidence $\alpha$ that entails $\mu$ given the
current belief state. (C4) simply says that no epistemic input can act
as its own defeater. Arlo-Costa and Parikh \cite{ap99} and
Lehmann \cite{leh95} have critically commented on (C2)
as have Freund and
Lehmann \cite{fl94} who have shown that it is inconsistent with the original AGM
axioms for belief sets (as is the weaker axiom, $C2^{'}$ proposed in Nayak {\it
et. al.} \cite{nfps96}). This last objection, as noted above, is no longer a
problem when the postulates are reformulated for epistemic states.

\section{The new recovery postulates}

In this section we provide additional recovery-like postulates and then
provide a semantic condition that provides the means with which to carry
out iterated revision. These additional properties
can be viewed as desirable properties for iterated
revision and cover a variety of situations, ranging from
sequences of revisions and contractions by the same formula to sequences
of revisions and contractions by a formula and its negation. In
particular these properties describe the conditions under which we can expect
stability or minimal loss of beliefs in the original epistemic state. Note
that in all of these properties the sequence of belief changes reverses
that in the original formulation of the recovery axiom where contraction
is followed by expansion. Stating the postulates in this form enables the connection
with iterated revision to become clear since it is in the case of revision followed by contraction that a notion of iterated revision is necessary (in the
original formulation of the recovery axiom, expansion is equivalent to revision thus obviating the need for a framework that requires iteration). In the postulates we make the implicit assumption
that information is received from the same source.

\begin{itemize}
\item (R1) $K(\Phi*\alpha-\alpha)\subseteq K(\Phi-\alpha)$
\item (R2) $\alpha,\lnot\alpha\notin K(\Phi)$ implies $K(\Phi)\subseteq K(\Phi*\alpha-\alpha)$
\item (R3) $\alpha\notin K(\Phi)$ implies $K(\Phi)\subseteq K(\Phi*\alpha*\lnot\alpha)$
\item (R4) $\alpha\in K(\Phi)$ implies  $K(\Phi-\alpha)\subseteq K(\Phi*\alpha-
\alpha)$

\end{itemize}

(R1) says that the result of revising an epistemic state and then
contracting by the same formula is always contained in the knowledge base
obtained after simply contracting by the same formula. (If I add the belief that
Cleopatra has children and then contract by this belief, the resultant knowledge base should be contained in the knowledge base obtained by my simply contracting by the belief that Cleopatra has children). (R2) says that
if neither a formula nor its negation are in the knowledge base associated
with an epistemic state then the original base will be contained in
that obtained after revision and contraction by the same formula.  (R3)
says that if a piece of information is not contained in the knowledge base
associated with an epistemic state, then a revision by that formula
followed by its negation will always include the original knowledge base.
(R4) says that if a formula is contained in the original knowledge base
then contracting by the same formula will produce a knowledge
base that is contained in one obtained by revising and contracting by the same formula.

The following additional properties further place conditions on recovery like situations:

\begin{itemize}
\item (R5) $K(\Phi*\alpha-\alpha)\subseteq K(\Phi)$
\item (R6) $\alpha\notin K(\Phi)$ implies $K(\Phi)\subseteq K(\Phi*\alpha-\alpha*\lnot\alpha)$
\item (R7) $\lnot\alpha\in K(\Phi)$ implies  $K(\Phi)\subseteq K(\Phi*\alpha-\alpha)$
\item (R8) $\alpha\in K(\Phi)$ implies  $K(\Phi)\subseteq K(\Phi*\alpha-\alpha*\alpha)$
\item (R9) $\alpha,\lnot\alpha\notin K(\Phi)$ implies $K(\Phi-\alpha)\subseteq K(\Phi*\alpha-\alpha)$
\end{itemize}

(R5) says that the knowledge base obtained by revising by an input and then contracting by it is contained
in the knowledge base associated with the original epistemic state.  (R6)
says that if a belief is not contained in the original knowledge base,
then the knowledge base is contained in the result of revising by a
formula, contracting it and then revising by its negation. (R7) says that
if a belief is not contained in the original knowledge base, then
the original knowledge base is contained in that obtained after revising and
contracting by its negation.
(R8) says that if a belief is contained in the original knowledge base,
then that belief will be preserved under a sequence of revisions which begin
with revision followed by contraction and then revision.
(R9) says that if the original knowledge base is agnostic about
a particular belief then contracting by that belief will result in a knowledge
base that is contained in one obtained by revising and then contracting
by that belief.
\begin{observation} \hspace{.00mm} \\
\begin{enumerate}
\item (R3) holds because $K(\Phi)$ is consistent.
\item (R9) follows from (R2).
\item (R5) follows from (R1) and the reformulated AGM postulates.
\item (R6) is the same as (R3), given the reformulated AGM postulates.
\item (R7) contradicts $(\Phi-2)$ and $(\Phi*2)$.
\item (R1) follows from (R5) if $\alpha\notin K(\Phi)$.
\item (R8) follows from the reformulated AGM postulates.
\end{enumerate}
\end{observation}

The reformulated AGM postulates for epistemic states and our additional recovery postulates, in our opinion, provide
a comprehensive framework for iterated revision which does justice
to the intuitions expressed in the original recovery axiom.
One of our stated aims is to link up $K(\Phi * \alpha - \alpha)$ and
$K(\Phi)$. We do this via (R2), (R4), (R5), (R7) and (R8). And it is (R7)
which contradicts the reformulated AGM postulates, as we have seen. Also, (R8) follows from AGM
anyway. Another way to put it: if $\alpha,\lnot\alpha\notin K(\Phi)$ then
$K(\Phi)=K(\Phi*\alpha-\alpha)$. If $\lnot\alpha\in K(\Phi)$ then AGM
prevents $K(\Phi)=K(\Phi*\alpha-\alpha)$. If $\alpha\in K(\Phi)$ then,
since $\alpha\notin K(\Phi*\alpha-\alpha)$ by AGM, it is AGM that prevents
$K(\Phi)=K(\Phi*\alpha-\alpha)$.

\subsection{Semantic properties}
We now provide conditions in semantic terms on revisions of epistemic states.
The following lay conditions on the positions of valuations by revision.

\begin{itemize}
\item  (S1) $M_{\preceq_{\Phi}}(\lnot\alpha)\subseteq
M_{\preceq_{\Phi*\alpha}}(\lnot\alpha)$
\medskip
\item (S2) $M_{\preceq_{\Phi*\alpha}}(\lnot\alpha)\subseteq
M_{\preceq_{\Phi}}(\lnot\alpha)$
\end{itemize}

The semantic properties taken together state an equality between the
minimal models of $\neg\alpha$ in the epistemic state prior to revision
and after revision. (S1) and (S2) taken together state that these minimal
models of $\neg\alpha$ retain their position after revision by $\alpha$.
For ease of statement of Theorem 1 below, we state these properties
as two separate containments rather than the implied equality.
The property stated here is straightforward.
Consider the minimal models of $\neg\alpha$ in the
total preorder associated with the epistemic state; these might or might
not be included in the minimal models of the total preorder itself. After revision by $\alpha$, the minimal models of the ordering cannot contain any
$\neg\alpha$ models. So the minimal models of $\neg\alpha$ are either demoted
in the ordering or stay where they are. Whatever be the case, no models of
$\neg\alpha$ can be promoted in the ordering to join the old minimal models
of $\neg\alpha$ and furthermore, none of the minimal models of $\neg\alpha$
are demoted. Revision by $\alpha$ can increase the plausibility
of $\alpha$ and decrease that of $\neg\alpha$; it certainly cannot increase
the plausibility of $\neg\alpha$.
Remarkably, this simple condition provides
all the semantic linkage we need with the numerous syntactic properties (R1-6, R8-9)
stated above.
It should be clear that the semantic properties stated above
are a weaker version of the (C2) postulate since in the Darwiche-Pearl framework,
which relies on a form of Spohnian conditioning \cite{Spohn:88a}, the position of {\it all} $\neg\alpha$ models is determined in the new epistemic state (via pointwise decrease
in their plausibility by one rank after revision by $\alpha$, thus preserving their relative ordering in the new epistemic state) whereas in our condition, we
simply specify the minimal models of $\neg\alpha$ in the new epistemic state.
Strengthening these postulates is possible, but possibly counterproductive and in any case, it is not our
present concern.

\begin{theorem}
Let $*$ and $\mdot$ be belief change operations on epistemic states
satisfying the reformulated AGM postulates.

\begin{enumerate}
\item $*$ and $\mdot$ satisfy (R1) iff $*$ satisfies (S1).
\item $*$ and $\mdot$ satisfy (R2)-(R4) iff $*$ satisfies (S2).
\end{enumerate}
\end{theorem}
\begin{proof}
\begin{enumerate}

\item (S1) follows immediately from (R1).  Suppose (S1) and pick a $u\in
M(K(\Phi-\alpha))$. If $u\in M(\alpha)$ then $u\in
M(K(\Phi*\alpha-\alpha))$ by AGM. If $u\in M(\lnot\alpha)$ then $u\in
M_{\preceq_{\Phi}}(\lnot\alpha)$. By (S1), $u\in
M_{\preceq_{\Phi*\alpha}}(\lnot\alpha)$. Therefore $u\in
M(K(\Phi*\alpha-\alpha))$.

\item Suppose (S2). Now suppose $\alpha,\lnot\alpha\notin K(\Phi)$. Pick a
$u\in M(K(\Phi*\alpha-\alpha))$. If $u\in M(\alpha)$ then $u\in
M(K(\Phi))$ by AGM. Otherwise $u\in M(K(\Phi))$ by (S2). So (R2) holds.
Now suppose $\alpha\notin K(\Phi)$. Pick a $u\in
M(K(\Phi*\alpha*\lnot\alpha))$. Since $u\in M(\lnot\alpha)$ it follows
that $u\in M(K(\Phi))$ by (S2). So (R3) holds. Now suppose $\alpha\in
K(\Phi)$. Pick a $u\in M(K(\Phi*\alpha-\alpha))$. If $u\in M(\alpha)$ then
$u\in M(K(\Phi))$ by AGM. Otherwise $u\in M(K(\Phi))$ by (S2). So (R4)
holds. Conversely, suppose (R2)-(R4).  If $\alpha,\lnot\alpha\notin
K(\Phi)$ then (S2) follows from (R2). If $\lnot\alpha\in K(\Phi)$ then
(S2) follows from (R3). If $\alpha\in K(\Phi)$ then (S2) follows from
(R4). \end{enumerate}

\end{proof}


The following shows that the case we were interested in, the relationship
between $K(\Phi*\alpha-\alpha)=K(\Phi-\alpha)$ is one of equality in
the case when $\neg\alpha$ is not contained in the original knowledge base.

\begin{corollary}
>From (R1)-(R4) it follows that, if $\lnot\alpha\notin K(\Phi)$ then
$K(\Phi*\alpha-\alpha)=K(\Phi-\alpha)$.
\end{corollary}
\begin{proof}
Follows from (S1) and (S2), which state together that
$M_{\preceq_{\Phi*\alpha}}(\lnot\alpha)=M_{\preceq_{\Phi}}(\lnot\alpha)$.
\end{proof}

Furthermore, note that since $\*$ and $\mdot$ are operations that satisfy
the reformulated AGM postulates, it follows that they satisfy (R5), (R6),
(R8) and (R9) as well.
\subsection{C2 and the new recovery postulates}
The connections between (S1), (S2) and (C2) are
interesting.
Objections to (C2) rely on the observation that
revising a belief state $\psi$ with a sentence of the form $p_{1} \wedge
p_{2} \wedge \ldots \wedge p_{n} \wedge q$ followed by a revision with
$\neg q$ reduces to revision with $\neg q$. Thus the (potentially useful)
belief in the conjunct $p_{1} \wedge p_{2} \wedge \ldots \wedge p_{n}$
is discarded (unless it was believed in the first place) even though
it does not in itself contradict $\neg q$. It can be argued that these
criticisms of the (C2) postulate are somewhat unfair, since this
unintuitive outcome does not follow if revision by $p_{1} \wedge
p_{2} \wedge \ldots \wedge p_{n} \wedge q$ is replaced by a sequence of
revisions by each of the conjuncts. One would revise with the full conjunction
only if these beliefs were somehow implicitly related. One scenario
where this behaviour required by the (C2) postulate appears to be fully
justified is when a source provides $p_{1} \wedge p_{2} \wedge
\ldots \wedge p_{n} \wedge q$ as an input, and subsequently changes its
mind (thus revising by $\neg q$). In a similar vein, if two consecutive
sensor readings contradict each other, it makes more sense to believe
the more recent reading, even if the previous reading provided
additional information. The (C2) postulate has also been criticized
from other perspectives. Cantwell \cite{Cantwell:99a} uses a version of
the George-the-criminal example to criticize the (C2) postulate. We note
that it is possible to argue against Cantwell's criticism along similar
lines to our arguments against the Cleopatra example (if the inputs come
from the same source, then the outcomes are intuitive, while inputs from
different sources would appear as distinct sentences, making the example
redundant).

The following example, a variation of
the George-the-criminal setting, makes clear that (C2) is too strong, and that (S1) and (S2)
are useful alternative weakenings. Assume that we start
by believing George is an armed robber. Our friend the police detective
tells us that this is incorrect, since no criminal records can be found
for George. Subsequently, she corrects her original
statement---she did find a criminal dossier on George at police
headquarters (it had been misplaced) and given its location, it
could have only come off the stack of files for people convicted of
illegal gun possession or the stack of convicted shoplifters' dossiers.
We must now revise our beliefs with the information that George is not an
armed robber, but either a shoplifter or a person convicted of illegal
gun possession. We construct below a scenario where the (C2) postulate
forces us to believe that George was convicted of illegal gun possession
(clearly too strong given the available evidence). We let $r$
denote `George is an armed robber', $g$ denote `George has been
convicted of illegal gun possession' and $s$ denote `George is a
convicted shoplifter' and use $c$ as an abbreviation for `George is a criminal' i.e.,
$r \vee g \vee  s$. Given the propositional language $\{r,
g, s\}$, we will represent models as sequences of 0's and 1's,
representing the valuations of $r$, $g$ and $s$ respectively (thus 100
represents a model in which $r$ is true and $g$ and $s$ are false). We assume for the sake of explanatory convenience that epistemic states
map valuations to natural numbers with the minimal models being identified
as those assigned the lowest rank (not necessarily 0)---thus inducing a total preorder on valuations.
Let the initial epistemic state $\Phi_{1}$ be defined as follows:\\

\noindent $\Phi_{1}(100) = \Phi_{1}(101) = \Phi_{1}(110) = \Phi_{1}(111) = 0$\\
\noindent $\Phi_{1}(010) = \Phi_{1}(011) = 1$\\
\noindent $\Phi_{1}(000) = \Phi_{1}(001) = 2$

Observe that, next to the models of $r$, we believe the
models of $g$ to be most plausible, reflecting the intuition that if
George is not an armed robber, then the next most likely scenario is
where George is in illegal possession of firearms.
To satisfy (C2) the epistemic state $\Phi_{2} = \Phi_{1} * \neg c$ must
appear as follows:\\

\noindent $\Phi_{2}(000) = 0$\\
\noindent $\Phi_{2}(100) = \Phi_{2}(101) = \Phi_{2}(110) = \Phi_{2}(111) = 1$\\
\noindent $\Phi_{2}(010) = \Phi_{2}(011) = 2$\\
\noindent $\Phi_{2}(001) = 3$

This leads to the epistemic state $\Phi_{3} = \Phi_{2} * \neg r
\wedge (g \vee s)$ where:\\

\noindent $\Phi_{3}(010) = \Phi_{3}(011) = 0$\\
\noindent $\Phi_{3}(000) = 1$\\
\noindent $\Phi_{3}(100) = \Phi_{3}(101) = \Phi_{3}(110) = \Phi_{3}(111) = 2$\\
\noindent $\Phi_{2}(001) = 3$

Observe that $g \in K(\Phi_{3})$, i.e., we are forced to believe George
has been convicted of illegal gun possession. If we relax (C2) with (S1)
and (S2), a permissible outcome of revising
$\Phi_{1}$ by $\neg c$ is the epistemic state $\Phi_{2}'$ where:\\

\noindent $\Phi_{2}'(000) = 0$\\
\noindent $\Phi_{2}'(100) = \Phi_{2}'(101) = \Phi_{2}'(110) = \Phi_{2}'(111) = 1$\\
\noindent $\Phi_{2}'(010) = \Phi_{2}'(011) = \Phi_{2}'(001) = 2$

Further revising with $\neg r \wedge (g \vee s)$ gives us the epistemic
state $\Phi_{3}'$ where:\\

\noindent $\Phi_{3}'(010) = \Phi_{3}'(011) = \Phi_{3}'(001) = 0$\\
\noindent $\Phi_{3}'(000) = 1$\\
\noindent $\Phi_{3}'(100) = \Phi_{3}'(101) = \Phi_{3}'(110) = \Phi_{3}'(111) = 2$

Notice that $g \not\in K(\Phi_{3})$.

\section{Conclusion}

In this paper we have shown how the intuitions underlying the axiom of
recovery can be rescued by paying attention to the assumptions underlying
putative counterexamples. We argued that the axiom of recovery places an
important rationality constraint on iterated revision, a framework that
requires that we think of revision as taking place on epistemic states
which encode preferences rather than just flat belief sets. We believe the
connection between the axiom of recovery and the (C2) postulate of
Darwiche-Pearl to be an interesting one.
For future work it
might be interesting to try and obtain a weakened version of the (C1)
postulate in a way that is similar to what we have done in this paper.

\bibliographystyle{ecai2002.bst}
\bibliography{thesis,master}
\end{document}